\documentclass[runningheads]{llncs}
\usepackage{graphicx}
\usepackage[noend]{algpseudocode}
\usepackage{algorithmicx,algorithm}
\usepackage{amsthm,amsmath,amssymb}
\usepackage{mathrsfs}
\usepackage[misc]{ifsym} 

\usepackage{multirow}
\usepackage{booktabs}
\usepackage{color}
\usepackage[colorlinks,linkcolor=blue]{hyperref}
\newcommand{\repeatthanks}{\textsuperscript{\thefootnote}}

\begin{document}
	\title{Self Context and Shape Prior for Sensorless Freehand 3D Ultrasound Reconstruction}
	\titlerunning{Self Context \& Shape Prior for Sensorless Freehand 3D US Reconstruction}
	%
	\author{
    Mingyuan Luo
    \inst{1,2,3}\thanks{Mingyuan Luo and Xin Yang contribute equally to this work.} \and
	Xin Yang
    \inst{1,2,3}\repeatthanks \and
	Xiaoqiong Huang
    \inst{1,2,3} \and
	Yuhao Huang
    \inst{1,2,3} \and
	Yuxin Zou
    \inst{1,2,3} \and
	Xindi Hu
    \inst{4} \and
	Nishant Ravikumar
    \inst{5,6} \and 
	Alejandro F Frangi
    \inst{1,5,6,7} \and
    Dong Ni
    \inst{1,2,3}\textsuperscript{(\Letter)}}


	\authorrunning{M. Luo et al.}
	%
	\institute{National-Regional Key Technology Engineering Laboratory for Medical Ultrasound, School of Biomedical Engineering, Health Science Center, Shenzhen University, China \\
	\email{nidong@szu.edu.cn} \and
	Medical Ultrasound Image Computing (MUSIC) Lab, Shenzhen University, China \and
	Marshall Laboratory of Biomedical Engineering, Shenzhen University, China \and
     School of Biomedical Engineering and Information, Nanjing Medical University, China \and
     Centre for Computational Imaging and Simulation Technologies in Biomedicine (CISTIB), University of Leeds, UK \and
     Leeds Institute of Cardiovascular and Metabolic Medicine, University of Leeds, UK \and
     Medical Imaging Research Center (MIRC), KU Leuven, Belgium
     }
	\maketitle              
	
	\begin{abstract}
    3D ultrasound (US) is widely used for its rich diagnostic information. However, it is criticized for its limited field of view. 3D freehand US reconstruction is promising in addressing the problem by providing broad range and freeform scan. The existing deep learning based methods only focus on the basic cases of skill sequences, and the model relies on the training data heavily. The sequences in real clinical practice are a mix of diverse skills and have complex scanning paths. Besides, deep models should adapt themselves to the testing cases with prior knowledge for better robustness, rather than only fit to the training cases.
    In this paper, we propose a novel approach to sensorless freehand 3D US reconstruction considering the complex skill sequences. Our contribution is three-fold. First, we advance a novel online learning framework by designing a differentiable reconstruction algorithm. It realizes an end-to-end optimization from section sequences to the reconstructed volume. 
    Second, a self-supervised learning method is developed to explore the context information that reconstructed by the testing data itself, promoting the perception of the model. 
    Third, inspired by the effectiveness of shape prior, we also introduce adversarial training to strengthen the learning of anatomical shape prior in the reconstructed volume. By mining the context and structural cues of the testing data, our online learning methods can drive the model to handle complex skill sequences. Experimental results on developmental dysplasia of the hip US and fetal US datasets show that, our proposed method can outperform the start-of-the-art methods regarding the shift errors and path similarities.
	\end{abstract}

	\section{Introduction}
\begin{figure}[t]
	    \centering
	    \includegraphics[width=0.9\textwidth]{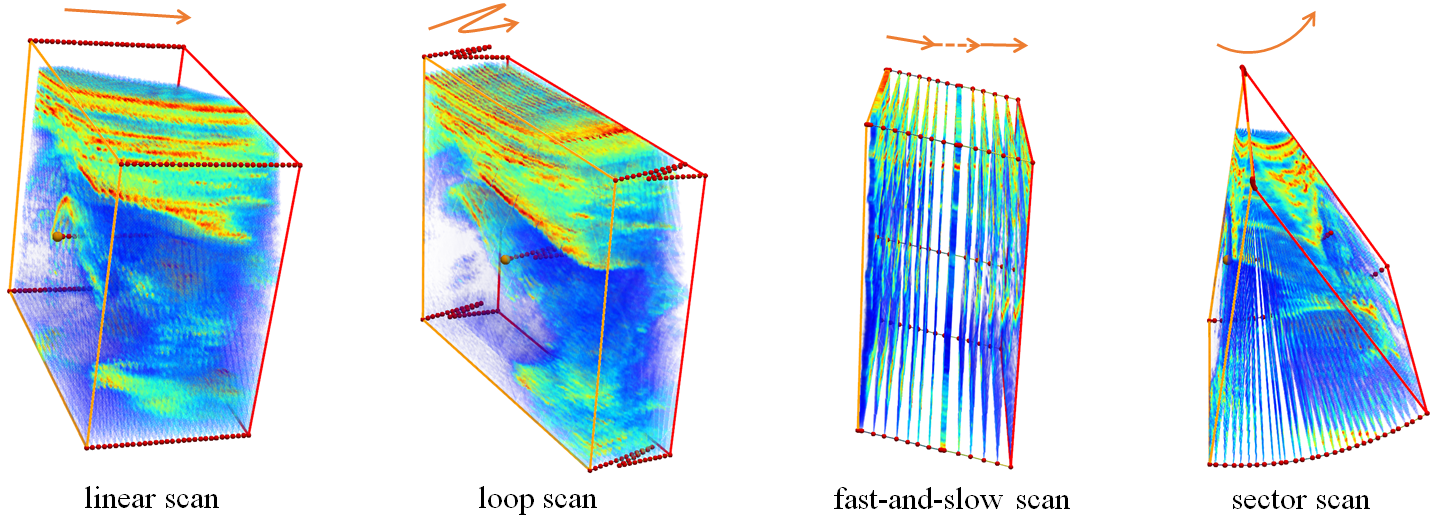}
	    \caption{Complex scan sequences considered in this work.}
	    \label{fig:1}
	\end{figure}

	Ultrasound (US) imaging is one of the main diagnostic tools in clinical due to its safety, portability, and low-cost. 3D US is increasingly used in clinical diagnosis~\cite{liu2019,looney2018fully,huang2020searching} because of its rich context informations which are not offered in 2D US. However, 3D US probe is constrained by the limited field of view and poor operability. Thus, exploring 3D freehand US reconstruction from a series of 2D frames has great application benefits~\cite{mohamed2019survey}. However, this kind of reconstruction is non-trivial due to the complex in-plane and out-plane shifts among adjacent frames, which are caused by the diverse scan skills and paths. In this paper, as the illustration in Fig.~\ref{fig:1}, besides the basic linear scan, we also consider three typical scan skills and their hybrid cases which are rarely handled in previous studies, including loop scan, fast-and-slow scan, and sector scan. \par	
	
	External sensor-based solutions can build tracking signal for 3D reconstruction~\cite{hennersperger2014vascular,lang2012multi}. But they are facing  magnetic influences, optical occlusion, and expensive costs. Therefore, US volume reconstruction from sensorless freehand scans plays an important role and has tremendous potential applications. 
Most previous researches are based on the non-deep learning methods~\cite{mercier2005,mozaffari2017}.
Prevost et al.~\cite{prevost2017deep} pioneered the deep learning based estimation of relative motion between US images, and later they extended their works by introducing the extra optical flow~\cite{farneback2003two} and integrating a sensor source~\cite{prevost20183d}. Guo et al.~\cite{guo2020sensorless} proposed a deep contextual learning network for reconstruction, which utilizes 3D convolutions over US video clips for feature extraction. 
However, these methods mainly handle the basic cases, rather than the complex scan sequences, such as the loop scan and fast-slow scan as shown in Fig.~\ref{fig:1}. In addition, previous models rely on the training data heavily and ignore the reconstruction robustness on the complex sequences during the test phase. 
Recently though, self-supervised learning (SSL) and adversarial learning (ADL) strategies in deep learning have been widely proven to learn an enriched representation from image itself and improve method robustness~\cite{yi2019,chen2019,hendrycks2019}. \par
	
	\begin{figure}[t]
		\centering
		\includegraphics[width=\textwidth]{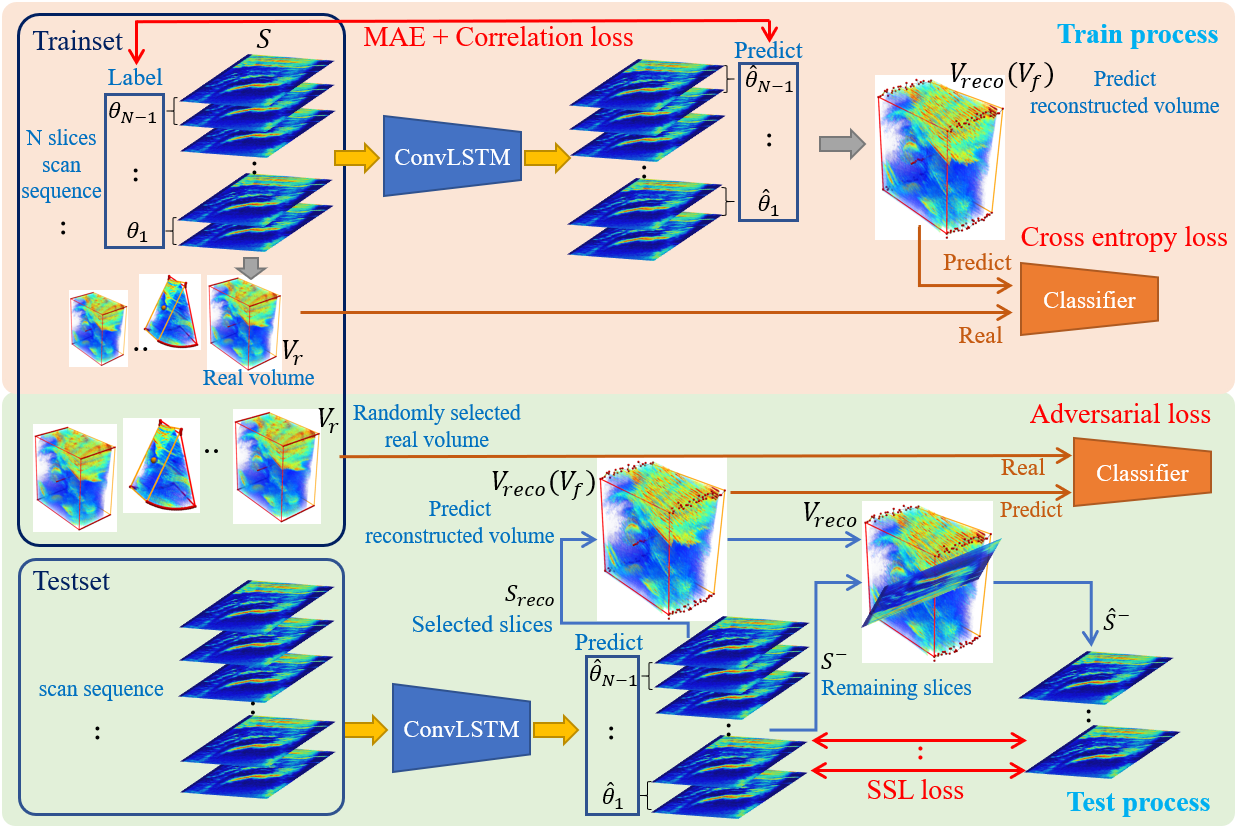}
		\caption{Overview of our proposed online learning framework.}
		\label{fig:2}
	\end{figure}
	
     For the limitations described above, we propose a novel approach to effectively address the difficulty of sensorless freehand 3D US reconstruction under complex skill sequences. 
	Our contribution is three-fold. First, a novel online learning framework is advanced by devising a differentiable algorithm to approximate the reconstruction process. It enables an end-to-end, recurrent optimization from section sequence to the whole reconstructed volume. Second, based on the differentiable design, a SSL scheme is developed to explore the context cues contained in the reconstructed volumes as pseudo supervisions, and hence regularize the prediction of future frames. Third, online ADL is introduced to strengthen the representation learning of anatomical shape priors, and thereby avoid irregular reconstructions. SSL and ADL in the online form are general. They can enable our method to generate plausible visual reconstructions when facing the difficult but practical scan cases that are shown in Fig.~\ref{fig:1}. \par
	
	\section{Methodology}
	Fig.~\ref{fig:2} shows the proposed online learning framework, including the recurrent Convolutional LSTM (ConvLSTM) backbone~\cite{shi2015convolutional} and the online learning framework to adapt to complex skill sequences. During the training process, the $N$-length sequence $S=\left\{s_i|i=1,\dots,N\right\}$ is input into the ConvLSTM to predict the 3D relative transformation parameters $\widehat{\Theta}=\{\widehat{\theta}_i|i=1,\dots,N-1\}$ among all adjacent frames, where $\widehat{\theta}_i$ (or $\theta_i$) represents a rigid transformation containing 3 translation and 3 rotation degrees. During the test, SSL and ADL are combined to form an online learning process to optimize the model’s estimation of complex skill sequences. \par
	
	\subsection{Online Self-Supervised Learning for Context Consistency}
	As shown in Fig.~\ref{fig:2}, for each frame in a sequence, its 3D transformation estimation aggregates the context of its neighboring frames. Therefore, there is an inherent context consistency constraint for each frame. When applying the estimated transformation of each frame to extract the slice from the volume reconstructed by the rest of the sequence frames, we should get a slice with very similar content as the frame itself. This is a typical self-supervision. Specifically, during the test phase, the estimated relative position $\widehat{p}_i$ of any frame $s_i$ with respect to the first frame $s_1$ is calculated according to the estimated relative transform parameter $\widehat{\Theta}$. In the scan sequence $S$, we firstly uniformly select a certain proportion (we set 0.5) of frames $S_{reco}\subsetneqq S$. A volume $V_{reco}$ is then reconstructed by using $S_{reco}$ and its corresponding estimated relative position $\widehat{P}_{reco}=\left\{\widehat{p}_i|s_i\in S_{reco}\right\}$ through a differentiable reconstruction approximation. For the remaining frames $S^{-}=S-S_{reco}$, we perform slice operation in $V_{reco}$ according to the corresponding estimated relative position $\widehat{P}^{-}=\left\{\widehat{p}_i|s_i\in S^{-}\right\}$ to get the generated slice $\widehat{S}^{-}$. The difference between $S^{-}$ and $\widehat{S}^{-}$ should be small to reflect the context consistency and is used to optimize the ConvLSTM. \par
	
	\subsection{Online Adversarial Learning for Shape Constraint}
	Shape prior is a strong constraint to regularize the volume reconstruction. Exposing the models to shape prior under the online learning scheme, rather than the offline training, provides us more chances to better explore the constraint for more generalizability, especially in handling the complex scan sequences where unplausible reconstructions often occur. Specifically, as shown in Fig.~\ref{fig:2}, we propose to encode the shape prior with an adversarial module.
The volume reconstruction $V_{r}$ from the training set is randomly selected as a sample with real anatomical structures. While the volume $V_{f}$ reconstructed from the testing sequence $S$ with relative estimated position parameters $\widehat{P}$ is taken as a sample of fake structure.
A classifier $C$ pre-trained to distinguish between the volume $V_{r}$ and $V_{f}$ reconstructed from all training sequences serves as the adversarial discriminator.
It adversarially tunes the structure contained in the reconstructed volume via the proposed differentiable reconstruction approximation. \par
	
	\subsection{Differentiable Reconstruction Approximation}
\begin{figure}[t]
	    \centering
	    \includegraphics[width=0.8\textwidth]{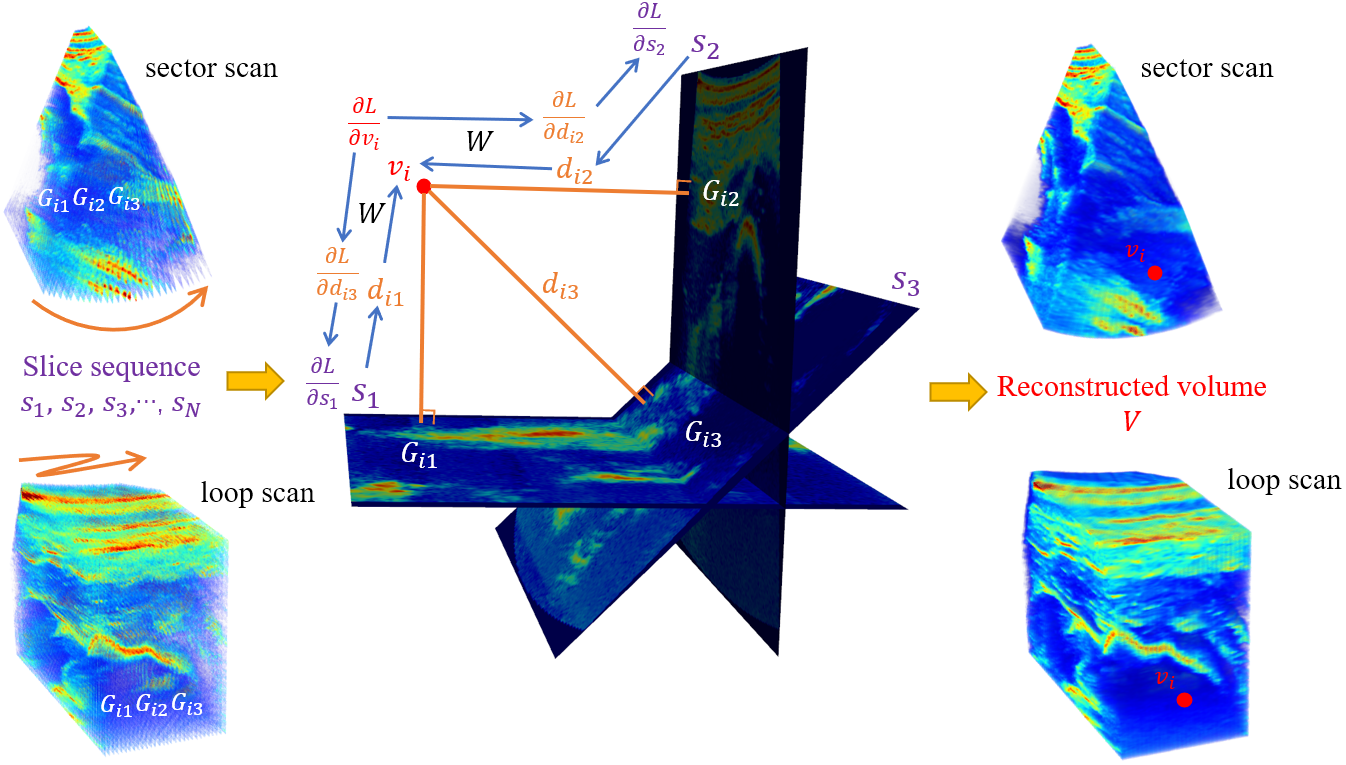}
	    \caption{Two reconstruction examples of differentiable reconstruction approximation.}
	    \label{fig:3}
	\end{figure}

	The end-to-end optimization of our online learning framework is inseparable from the differentiable reconstruction. In general, most of the unstructured interpolation operations in the reconstruction algorithm (such as the Delaunay algorithm) are not differentiable and make our online learning blocked. As a foundation of our online design, we firstly propose a differentiable reconstruction approximation to mimic the interpolation. As shown in Fig.~\ref{fig:3}, the volume $V$ is reconstructed using the $N$ slices $\left\{s_j|j=1,\dots,N\right\}$. For any pixel $v_i\in V$, the distance $d_{ij}$ from any slice $s_j$ and the gray value $G_{ij}$ of the projection points on any slice $s_j$ are calculated. Then the reconstructed gray value $G_{v_i}$ at pixel $v_i$ is calculated as: 
	\begin{equation}
	    G_{v_i}=\sum_{j}{W(d_{ij})G_{ij}}
	\end{equation}
	Among them, $W\left(\cdot\right)$ is a weight function. Its purpose is to encourage small $d_{ij}$. The core formulation is softmax operation on the reciprocal of $d_{ij}$:
	\begin{equation}
	    W(d_{ij})=\frac{\exp(1/(d_{ij}+\epsilon))}{\sum_j{\exp(1/(d_{ij}+\epsilon}))},
	\end{equation}
	where $\epsilon$ prevents division by $0$. If only the nearest slice is weighted, the approximation can be expressed as follows:
	\begin{equation}
	    W(d_{ij})=\frac{1}{Z}(\frac{1}{d_{ij}+\epsilon}\frac{\exp(1/(d_{ij}+\epsilon))}{\sum_j{\exp(1/(d_{ij}+\epsilon}))}),
	\end{equation}
	where $1/Z$ is the normalization coefficient such that $\sum_jW(d_{ij})=1$.
	
	Fig.~\ref{fig:3} shows two reconstruction examples. The left volume of each example directly puts the slice into the volume according to the position without interpolation, and the right is the result of differentiable reconstruction approximation.
	
	\subsection{Loss Function}
	In the training phase, the ConvLSTM predicts relative transform parameters $\widehat{\Theta}$ between all adjacent frames in the skill sequence. Its loss function includes two items, the first item is the mean absolute error (MAE) loss (i.e., L1 normalization), and the second item is the case-wise correlation loss from~\cite{guo2020sensorless}, which is beneficial to improve the generalization performance.
	\begin{equation}
	    L=\|\widehat{\Theta}-\Theta\|_1 + (1 - \frac{\textbf{Cov}(\widehat{\Theta},\Theta)}{\sigma(\widehat{\Theta})\sigma(\Theta)})
	\end{equation}
	where $\left\|\cdot\right\|_1$ indicates L1 normalization, $\textbf{Cov}$ gives the covariance, and $\sigma$ calculates the standard deviation. The classifier $C$ uses cross-entropy loss to distinguish whether the input volume is reconstructed or real volume. 
	
	In the test phase, the SSL and the ADL are jointly iteratively optimized. The first two items of $L_d$ are the adversarial loss used to optimize the classifier $C$, and the third item is the quadratic potential divergence from~\cite{su2018ganqp}, which helps to prevent the gradient vanishing without additional Lipschitz constraint. The first term of $L_g$ is the adversarial loss used to optimize the ConvLSTM, and the second term is the self-supervised loss.
	\begin{equation}
	    L_d=\mathbb{E}_{V_f\sim\mathbb{P}_{V_f},V_r\sim\mathbb{P}_{V_r}}[C(V_f)-C(V_r)+\frac{\|C(V_f)-C(V_r)\|_2^2}{2\|V_f-V_r\|_1}]
	\end{equation}
	\begin{equation}
	    L_g=-\mathbb{E}_{V_f\sim\mathbb{P}_{V_f}}[C(V_f)]+\|\widehat{S}^{-}-S^{-}\|_1
	\end{equation}
	where $\left\|\cdot\right\|_1$ and $\left\|\cdot\right\|_2$ indicate L1 and L2 normalization, respectively. \par
	
	\section{Experiments}
    \paragraph{\textbf{Materials and Implementation.}} Our experiments involve two datasets, i.e., 3D developmental dysplasia of the hip (DDH) US dataset and 3D fetus US dataset. The DDH dataset contains 101 US volumes from 14 volunteers. Its average volume size is $370\times403\times398$ with voxel resolution $0.1\times0.1\times0.1$ mm. The fetal dataset contains 78 fetal US volumes from 78 volunteers. The gestational age ranges from 10 to 14 weeks. Its average volume size is $402\times535\times276$, the voxel resolution is $0.3\times0.3\times0.3$ mm. All data collection is anonymous. The collection and use of data are approved by the local IRB. \par
	
	It is time-consuming and expensive to collect a large number of real freehand sequences with complex scan skills. In order to verify the proposed framework on the complex skill sequences, the real 3D volumes are used to generate massive complex skill sequences for our training and evaluation. Multiple complex scan sequences are dynamically simulated with diverse scan skills from each US volume to form our final corpus. Specifically, the scan sequences are a complex combination of loop scan, fast-and-slow scan and sector scan with the aim to simulate the loop movement, uneven speed, and anisotropic movement of probe. \par
	
	For the 3D DDH and fetal US dataset, we randomly divide the dataset into 85/15 and 65/13 volumes for training/testing according to the patients, and the sequence length is 120 and 90, respectively. The number of generated sequences is 100 and 10 for training and testing, respectively. All generated slices are with size $300\times300$ pixel. In the training phase, the Adam optimizer is used to iteratively optimize our ConvLSTM. The epoch and batchsize are 200 and 4, respectively. The learning rate is $10^{-3}$, and the learning rate is reduced by half every 30 epochs. The ConvLSTM has 1 layer and convolutional kernel size is 3. For classifier $C$, the epoch, batchsize and learning rate is 50, 1 and $10^{-4}$, respectively. During the test phase, for each testing sequence, we iterate the online learning with adversarial and self-supervised losses for 30 iterations, and the learning rate is set as $10^{-6}$. All codes are implemented in PyTorch~\cite{pytorch}.
	
	\paragraph{\textbf{Quantitative and Qualitative Analysis.}} The current commonly used evaluation indicator is the final drift~\cite{prevost20183d,guo2020sensorless}, which is the drift of final frame of a sequence, and the drift is the distance between the center points of the frame according to the real relative position and the estimated relative position. On this basis, a series of indicators are used to evaluate the accuracy of our proposed framework in estimating the relative transform parameters among adjacent frames. Final drift rate (FDR) is the final drift divided by the sequence length. Average drift rate (ADR) is the cumulative drift of all frames divided by the length from the frame to the starting point of the sequence, and finally, the average value is calculated. Maximum drift (MD) is the maximum accumulated drift of all frames. Sum of drift (SD) is the sum of accumulated drift of all frames. The bidirectional Hausdorff distance (HD) emphasizes the worst distances between the predicted positions (accumulatively calculated by the relative transform parameters) and the real positions of all the frames in the sequence.
	
	\begin{table}[t]
    	\centering
    	\caption{The mean (std) results of different methods on the DDH sequences.}
    	\scalebox{0.9}{
    	\begin{tabular}{l|c|c|c|c|c}
    		\toprule
    		\textbf{Methods}                 & \textbf{FDR(\%)} & \textbf{ADR(\%)} & \textbf{MD(mm)} & \textbf{SD(mm)} & \textbf{HD(mm)} \\ 
    		\hline
    		\hline
    		CNN~\cite{prevost20183d}  & 16.25(10.45) & 44.58(8.71) & 21.14(4.11) & 1798.71(433.18) & 21.10(4.10)     \\ \hline
    		DCL-Net~\cite{guo2020sensorless} & 12.78(14.87) & 25.46(23.48) & 9.07(7.99) & 363.68(345.06) & 7.43(5.43) \\ \hline
    		ConvLSTM       & 11.15(3.97) & 22.08(9.03) & 7.61(2.99) & 498.84(197.32) & 6.63(2.77) \\ \hline
    		ConvLSTM+SSL   & 10.89(4.05) & 21.93(9.03) & 7.52(2.94) & 493.44(197.70) & 6.58(2.76) \\ \hline
    		ConvLSTM+ADL   & 6.21(3.09)  & 13.89(6.95) & 4.70(2.30) & 291.01(142.08) & 4.10(1.88) \\ \hline
    		ConvLSTM+SSL+ADL & \textcolor{blue}{5.44(3.03)}  & \textcolor{blue}{13.47(6.47)} & \textcolor{blue}{4.44(2.29)} & \textcolor{blue}{274.14(136.89)} & \textcolor{blue}{3.91(1.89)} \\ 
    		\bottomrule
    	\end{tabular}}
    	\label{tab:1}
    \end{table}

	\begin{table}[t]
    	\centering
    	\caption{The mean (std) results of different methods on the fetus sequences.}
    	\scalebox{0.9}{
    	\begin{tabular}{l|c|c|c|c|c}
    		\toprule
    		\textbf{Methods}                 & \textbf{FDR(\%)} & \textbf{ADR(\%)} & \textbf{MD(mm)} & \textbf{SD(mm)} & \textbf{HD(mm)} \\ 
    		\hline
    		\hline
    		CNN~\cite{prevost20183d}         & 16.59(9.67) & 31.82(14.02) & 24.50(10.43) & 1432.89(698.88) & 19.16(8.21)      \\ \hline
    		DCL-Net~\cite{guo2020sensorless} & 12.47(8.49) & 30.58(16.05) & 17.74(8.56) & 882.00(452.00) & 15.88(7.71)      \\ \hline
    		ConvLSTM                         & 16.71(5.22) & 31.33(12.68) & 20.57(8.33) & 938.42(391.39) & 15.79(6.78) \\ \hline
    		ConvLSTM+SSL                     & 15.86(5.10) & 30.78(12.20) & 20.01(8.27) & 909.95(382.43) & 15.59(6.53) \\ \hline
    		ConvLSTM+ADL                     & 10.49(4.86) & 28.15(9.86) & 17.49(7.68) & 762.79(292.92) & 14.18(5.99) \\ \hline
    		ConvLSTM+SSL+ADL            & \textcolor{blue}{9.94(4.41)} &  \textcolor{blue}{27.08(9.31)} & \textcolor{blue}{16.84(7.56)} & \textcolor{blue}{730.11(303.20)} & \textcolor{blue}{14.12(5.89)}  \\ \bottomrule
    	\end{tabular}}
    	\label{tab:2}
    \end{table}
	
	Table~\ref{tab:1} \&~\ref{tab:2} summarize the overall comparison of the proposed online learning framework with other existing methods and ablation frameworks on simulated DDH or fetus complex skill sequences, respectively. “CNN”~\cite{prevost20183d} and “DCL-Net”~\cite{guo2020sensorless} are considered in this study for comparison. “SSL” and “ADL” are our proposed online self-supervised learning and adversarial learning method, respectively.

     It should be noted that we have updated the results of “DCL-Net” in the MICCAI version. Considering that the mean parameter in the “DCL-Net” setting is not suitable for scan sequences with large path variations, we removed the mean parameter in the updated experiments and only used the relative transformation parameters between all adjacent frames as the ground truth to obtain the final results. Compared with “CNN”, all indicators of “DCL-Net” have been significantly improved.

     As can be seen from Table~\ref{tab:1} \&~\ref{tab:2}, our proposed methods get consistent and better improvements both on DDH and fetal complex skill sequences. The comparison with the ablation frameworks fully proves the necessity of introducing online learning in the test phase. In particular, our proposed online learning methods help the ConvLSTM improve the accuracy by 5.71\%/2.72mm for DDH sequences, and 6.77\%/1.67mm for fetal sequences in terms of the FDR/HD indicators. Under the complementary effects of mining context consistency and shape constraint, the proposed framework has significantly robust the estimation accuracy of complex skill sequences.

	\begin{figure}[t]
	    \centering
	    \includegraphics[width=\textwidth]{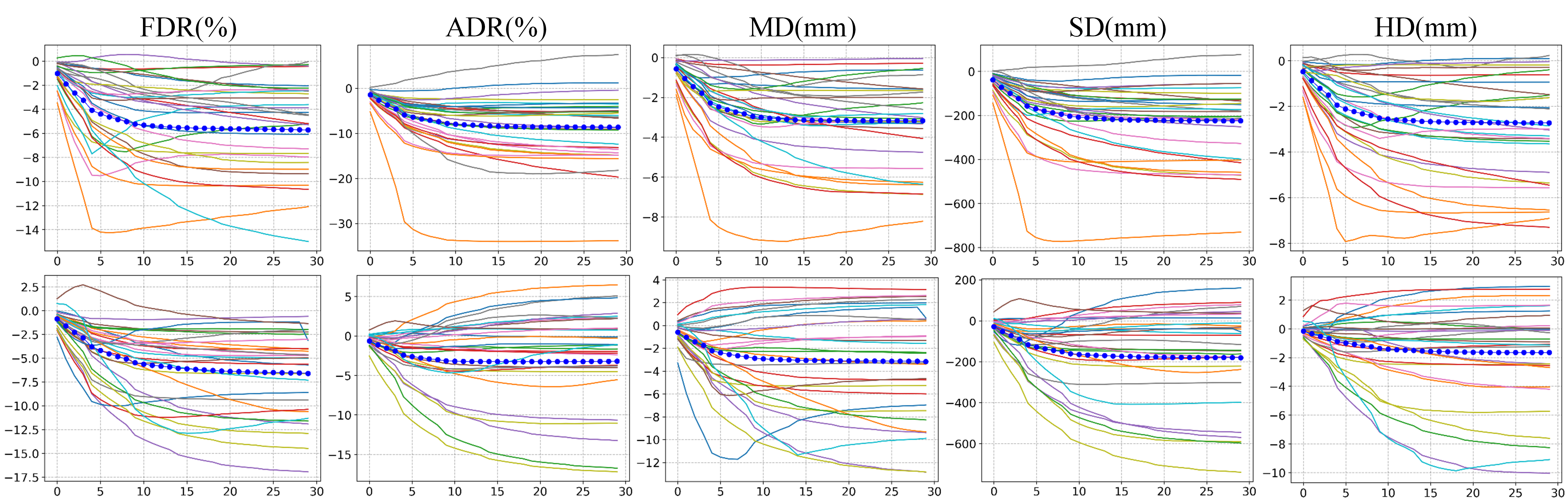}
	    \caption{Indicator declining curves for online iterative optimization on DDH (the first row) and fetus (the second row) US datasets.}
	    \label{fig:4}
	\end{figure}
	
	\begin{figure}[t]
	    \centering
	    \includegraphics[width=\textwidth]{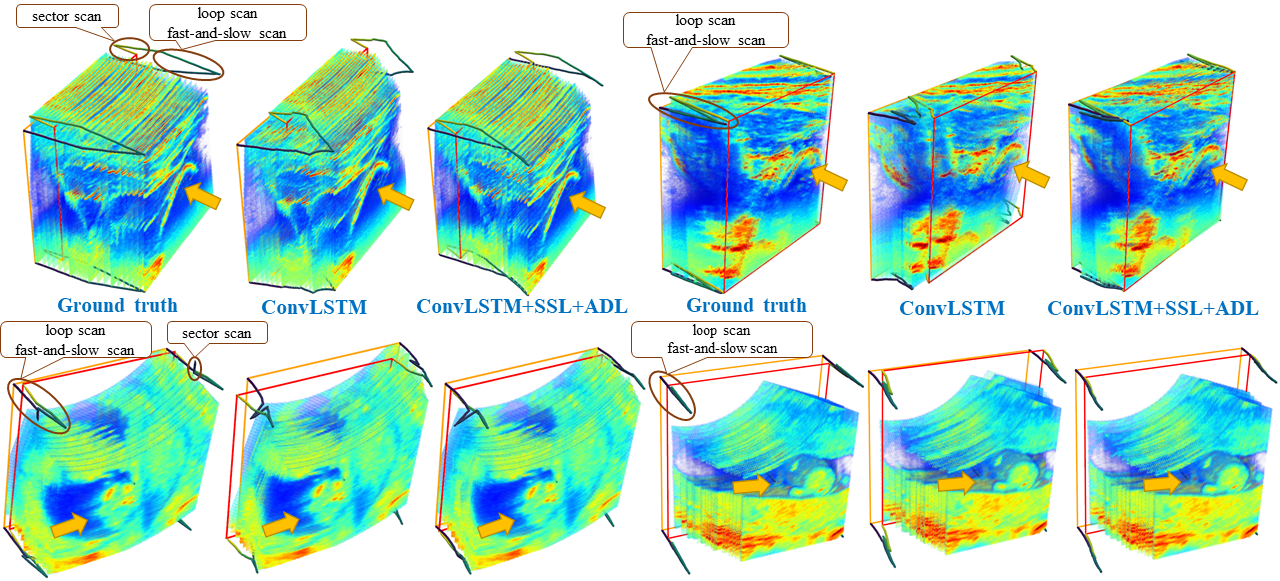}
	    \caption{Specific case of online iterative optimization. The first row and second row shows the DDH and fetus cases, and the yellow arrows indicate the anatomic structure.}
	    \label{fig:5}
	\end{figure}
	
	Fig.~\ref{fig:4} shows the indicator declining curves for online iterative optimization of the ConvLSTM on DDH and fetus US datasets. In each subfigure of Fig.~\ref{fig:4}, the abscissa and ordinate represent the number of iterations and the extent of indicators decline, respectively. The blue dot curve is the average indicator declining curve of all data. It is obvious that the indicators curve of almost all single sequences is declining significantly, and the downward trend is from rapid decline to gentle convergence. Among the various indicators, FDR has the largest decline, which is reduced to $48.78\%$ and $59.48\%$ of the ConvLSTM on the DDH and fetus US dataset, respectively. 
	Fig.~\ref{fig:5} shows four specific cases of online iterative optimization on DDH and fetus complex skill sequences, respectively. In each subfigure of Fig.~\ref{fig:5}, the orange and red box represent the first and final frame respectively and the pipeline that change color represent the scan path. It can be observed that through online iterative optimization, the final reconstruction result is significantly improved compared with the ConvLSTM, with smaller drift and more stable trajectory estimation.

	\section{Conclusion}
	In this paper, we proposed the first research about freehand complex scan 3D US volume reconstruction. Benefiting from self context, shape prior and differentiable reconstruction approximation, the proposed online learning framework can effectively solve the challenge of 3D ultrasound reconstruction under complex skill sequences. Experiments on two different US datasets prove the effectiveness and versatility of the proposed framework. Future research will focus on extending this framework to more scanning targets and even different modalities.

\subsubsection*{Acknowledgements}
This work was supported by the National Key R\&D Program of China (No. 2019YFC0118300), Shenzhen Peacock Plan (No. KQTD20160\\-53112051497, KQJSCX20180328095606003), Royal Academy of Engineering under the RAEng Chair in Emerging Technologies (CiET1919/19) scheme, EPSRC TUSCA (EP/V04799X/1) and the Royal Society CROSSLINK Exchange Programme (IES/NSFC/201380).
	
	%
	%
	%
	\bibliographystyle{splncs04}
	\bibliography{paper662}

\end{document}